\documentclass[]{segabs}

\usepackage{amssymb,amsmath}
\usepackage[utf8]{inputenc}
\usepackage{breqn}
\usepackage{graphicx}
\usepackage[colorlinks=true,urlcolor=blue,citecolor=black,linkcolor=black]{hyperref}
\usepackage{multirow}
\usepackage{wrapfig}

\newcommand{\B}{\mathbf}

\righthead{Memorization in learned priors for geophysical inversion}

\title{On the role of memorization in learned priors for
geophysical inverse problems}
\author{Ali Siahkoohi and Davide Sabeddu\\
University of Central Florida}

\begin{document}
\maketitle
\begin{abstract}
Learned priors based on deep generative models offer data-driven
regularization for seismic inversion, but training them requires a
dataset of representative subsurface models---a resource that is
inherently scarce in geoscience applications. Since the training objective of most generative models can be cast as
maximum likelihood on a finite dataset, any such model risks converging
to the empirical distribution---effectively memorizing the training
examples rather than learning the underlying geological distribution. We
show that the posterior under such a memorized prior reduces to a
reweighted empirical distribution---i.e., a likelihood-weighted
lookup among the stored training examples. For diffusion models
specifically, memorization yields a
Gaussian mixture prior in closed form, and linearizing the forward
operator around each training example gives a Gaussian mixture
posterior whose components have widths and shifts governed by the local
Jacobian. We validate these predictions on a stylized inverse problem
and demonstrate the consequences of memorization through
diffusion posterior sampling for full waveform
inversion. Code to partially reproduce the results is available at
\href{https://github.com/luqigroup/mempost}{github.com/luqigroup/mempost}.
\end{abstract}

\section{Introduction}

Seismic inversion is an ill-posed inverse problem whose solution
nonuniqueness and sensitivity to data noise make uncertainty
quantification essential for reliable subsurface characterization
\citep{tarantola2005inverse, stuart2010inverse}. A Bayesian framework addresses this need by
representing the solution as a posterior distribution, where the
prior encodes knowledge about plausible subsurface models. This
formulation places the quality of the prior at the center of the
inference \citep{hosseini2025error}.
Conventional priors---e.g., Gaussian or sparsity-promoting Laplace distributions---offer analytical
convenience but often fail to capture the complex structure
of realistic geological models.

To address this limitation, a growing body of work replaces
handcrafted priors with learned alternatives based on generative
models
\citep{mosser2020stochastic, zhang2021bayesian, siahkoohi2021deep,
zhao2022normalizing, zhang2023bayesian3d, yin2024wise,
zhao2025efficient, siahkoohi2026dualspace}. Among these, score-based
diffusion models \citep{song2019generative, ho2020denoising}---a class
of generative models that learn to reverse a gradual noising process in
order to draw samples from a target distribution---have received
considerable attention
\citep{erdinc2025power, ravasi2025measurement, zeng2026fwi}. Once
trained, the learned score function can guide posterior sampling
algorithms
\citep{chung2023diffusion, song2023pseudoinverse}, providing data-driven
regularization that reflects the statistical structure of geological
model datasets.

While these approaches have shown promise in controlled settings,
deploying learned diffusion priors for geoscience inverse problems
introduces a fundamental challenge: high-fidelity subsurface models
derived from well logs or high-resolution imaging are inherently
scarce, and synthetic alternatives based on geostatistical simulations
are often overly simplistic relative to real geological complexity. Augmenting the dataset with
samples synthesized by a generative model trained on the same limited
data is equally ineffective: iteratively retraining on
model-generated data leads to progressive quality
degradation---a phenomenon known as model autophagy
\citep{alemohammad2024mad}. As a result, the ratio of available training examples to model
dimensionality remains small, raising the risk of
memorization---i.e., the generative model reproduces training examples rather than
learning the underlying geological distribution. Specifically, in the context of diffusion models,
\citet{baptista2025memorization} show that the minimizer of the
empirical denoising score matching loss is the score of a Gaussian
mixture centered on the training examples, and the number of
training samples required to avoid memorization grows exponentially
with data dimensionality
\citep{biroli2024dynamical}---a threshold unlikely to be met for
high-dimensional problems. Since prior misspecification
propagates directly into the posterior
\citep{hosseini2025error}, understanding the consequences of
memorization for posterior inference is essential.

In this work, we characterize these consequences for geophysical
inverse problems. Our contributions are threefold: (i) we show that
when a generative model memorizes its training data, the posterior
reduces to a lookup-table---i.e., a likelihood-weighted selection
among stored training examples; (ii) for diffusion models, the
memorized prior takes the form of a Gaussian mixture, and
linearizing the forward operator around each training example yields
a closed-form Gaussian mixture posterior whose component means are
shifted by the adjoint Jacobian and whose covariances balance prior
constraint against data sensitivity; and (iii) we validate these
predictions analytically on a stylized inverse problem and
demonstrate the practical consequences through diffusion posterior
sampling for frequency-domain full waveform inversion (FWI). In the
following sections, we formalize the Bayesian inverse problem with a
learned prior, characterize the failure mode of memorization, derive
the linearized posterior in closed form, and conclude with numerical
illustrations.

\section{Inverse problems with learned priors}

We are concerned with Bayesian inference for ill-posed inverse problems
in which the prior distribution is learned from a finite dataset.
Consider a nonlinear forward operator
$\B{F}: \mathbb{R}^d \to \mathbb{R}^m$ relating the model parameters
$\B{x}$ to observed data $\B{y}$ via
$\B{y} = \B{F}(\B{x}) + \boldsymbol{\eta}$, where
$\boldsymbol{\eta} \sim \mathcal{N}(\B{0},\, \gamma^2 \B{I})$ and
$\gamma^2$ is the noise variance. In FWI, for instance, $\B{x}$ is
the subsurface velocity model and
$\B{F}$ encapsulates the solution of the wave equation together with
the observation operator. The Bayesian solution is the posterior
distribution obtained via Bayes' rule:
\begin{equation}
p(\B{x} \mid \B{y}) \propto
\underbrace{p(\B{y} \mid \B{x})}_{\text{likelihood}}\;
\underbrace{p(\B{x})}_{\text{prior}},
\label{eq:bayes}
\end{equation}
where the likelihood is
$p(\B{y} \mid \B{x}) = \mathcal{N}(\B{y} \mid
\B{F}(\B{x}),\, \gamma^2 \B{I})$.
Learned priors replace handcrafted distributional assumptions with a
parametric model $p_\theta(\B{x})$ trained on a dataset
$\{\B{x}_1, \ldots, \B{x}_N\}$ by maximum likelihood estimation.
Since the true data distribution
$p_{\mathrm{data}}$ is accessible only through the empirical
distribution $\hat{p}_N = \frac{1}{N}\sum_{n=1}^{N}
\delta(\B{x} - \B{x}_n)$, the quality of the learned prior
depends on whether the model generalizes beyond the training examples.
When it does not---i.e., when $p_\theta$ converges to
$\hat{p}_N$---we show below that the posterior inherits the discrete
structure of the empirical distribution.

\subsection{The empirical prior and the lookup-table posterior}

Most deep generative models used as learned priors---including
normalizing flows \citep{rezende2015variational}, variational
autoencoders \citep{kingma2014autoencoding}, and diffusion models
\citep{ho2020denoising, song2021scorebased}---are trained by
maximum likelihood estimation or a lower bound thereof, both
equivalent to minimizing a Kullback--Leibler divergence between
the data and model distributions \citep{kingma2023understanding}.
For a sufficiently expressive model class and without sufficient
regularization---whether explicit or through architectural inductive
bias---the Kullback--Leibler divergence between $\hat{p}_N$ and
$p_\theta$ is minimized when the two distributions coincide,
yielding $p_\theta(\B{x}) = \hat{p}_N(\B{x}) = \frac{1}{N}
\sum_{n=1}^{N} \delta(\B{x} - \B{x}_n)$. Substituting this memorized
prior into Equation~\ref{eq:bayes} yields
\begin{equation}
p(\B{x} \mid \B{y}) = \sum_{n=1}^{N} w_n\,
\delta(\B{x} - \B{x}_n),\
w_n \propto \exp\!\left(-\frac{\|\B{F}(\B{x}_n) -
\B{y}\|_2^2}{2\gamma^2}\right).
\label{eq:lookup-table}
\end{equation}
The posterior is a lookup table: discrete probability mass on each
training example, weighted by data fit. The resulting uncertainty
reflects which stored models best explain the data, not 
geological variability. This failure mode applies to any generative model trained with maximum likelihood estimation.

\subsection{Memorized diffusion models as Gaussian mixture priors}

For diffusion models, in particular,
\citet{baptista2025memorization} establish a concrete characterization
of memorization. The forward diffusion process maps each data sample
$\B{x}_0$ to
$\B{x}_t \sim \mathcal{N}(m(t)\,\B{x}_0,\, \sigma^2(t)\,\B{I})$,
where $m(t)$ and $\sigma(t)$ control the signal attenuation and noise
level at each step of the forward process. The score
matching loss trains the model to approximate the score
function---i.e., the gradient of the log-density---of the noisy data
distribution at each $t$; when minimized exactly on a finite dataset,
the minimizer is the score of a Gaussian mixture:
\begin{equation}
p^N\!(\B{x}, t) = \frac{1}{N} \sum_{n=1}^{N}
\mathcal{N}\!\left(\B{x} \mid m(t)\,\B{x}_n,\, \sigma^2(t)\,
\B{I}\right).
\label{eq:gmm-prior}
\end{equation}
At large $t$, the components overlap and the mixture approximates a
smooth density; as $t \to 0$, $m(t) \to 1$ and $\sigma(t) \to 0$,
and the mixture recovers the empirical distribution $\hat{p}_N$.
While the implicit regularization of overparameterized neural
networks can delay convergence to this exact minimizer, the number
of training samples required to avoid memorization grows
exponentially with the data dimensionality
\citep{biroli2024dynamical, bonnaire2025why}---placing
high-dimensional geoscience applications squarely in the memorization
regime.

\begin{figure*}[!t]
\centering
\setlength{\tabcolsep}{2pt}
\renewcommand{\arraystretch}{0}
\begin{tabular}{@{}c@{}c@{}c@{}}
\includegraphics[width=0.31\textwidth]{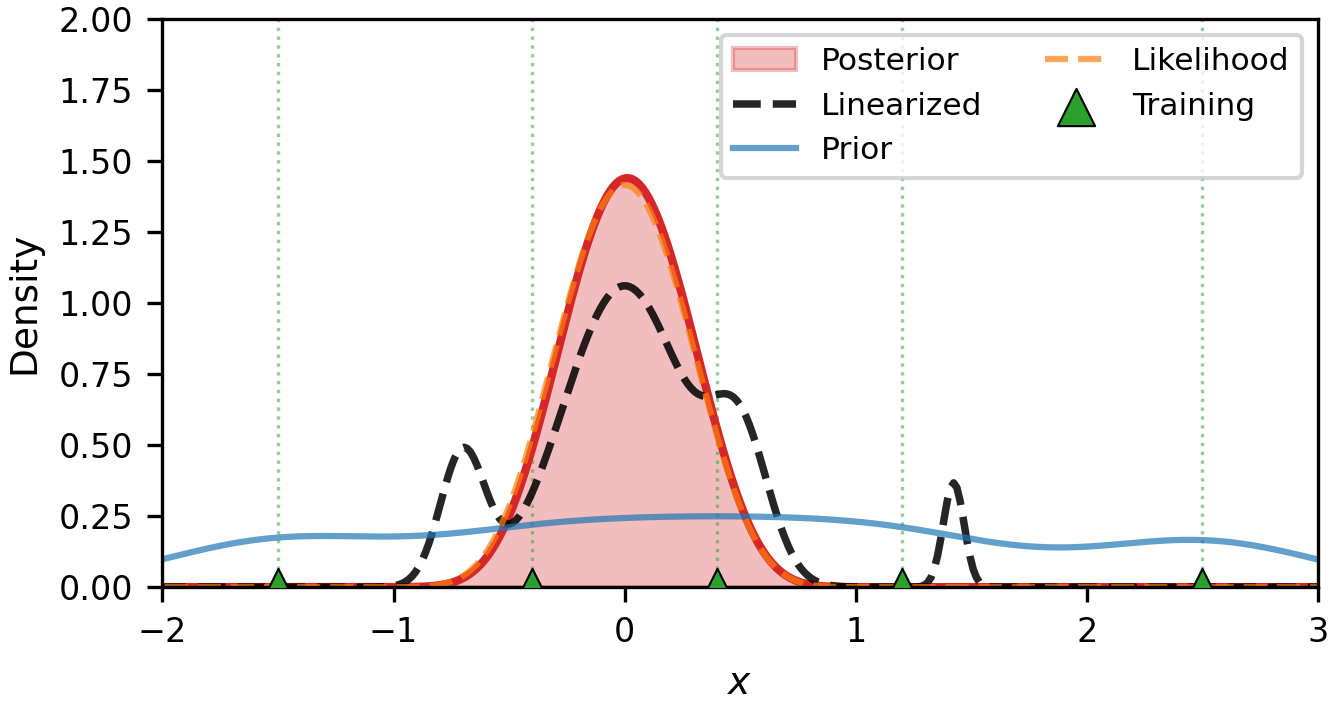} &
\includegraphics[width=0.31\textwidth]{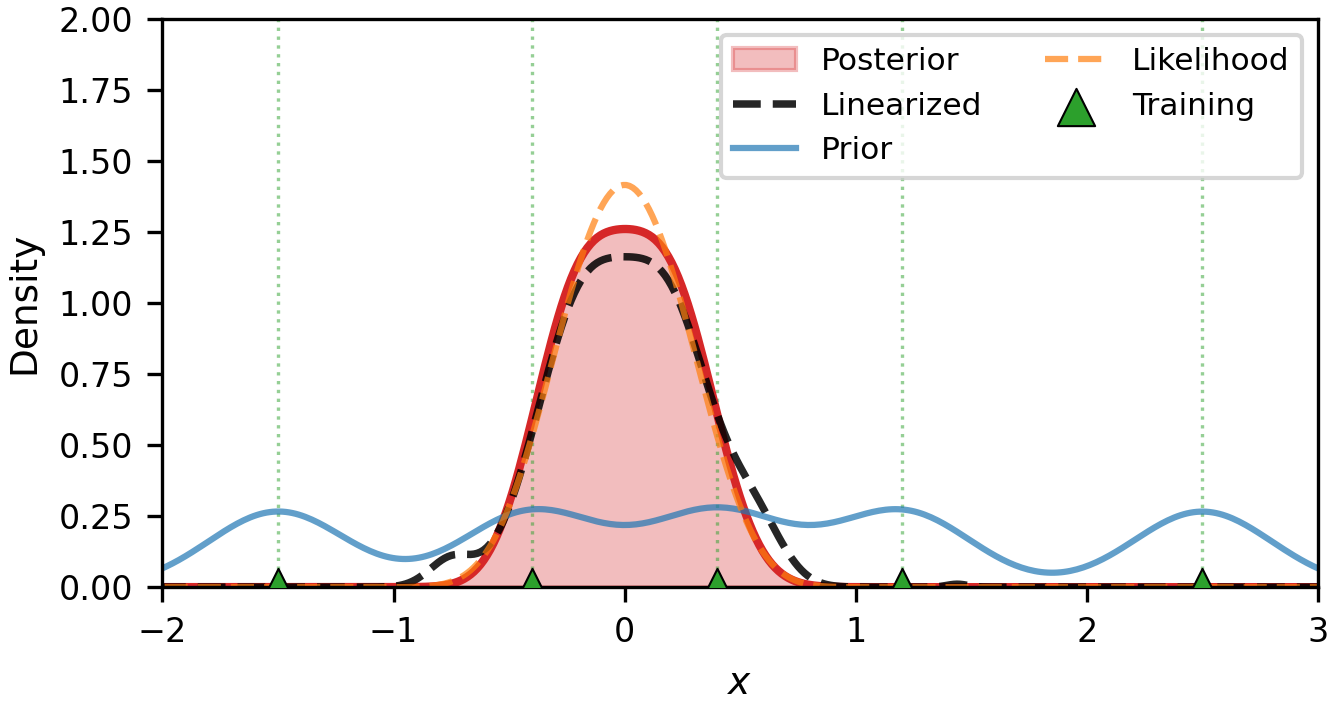} &
\includegraphics[width=0.31\textwidth]{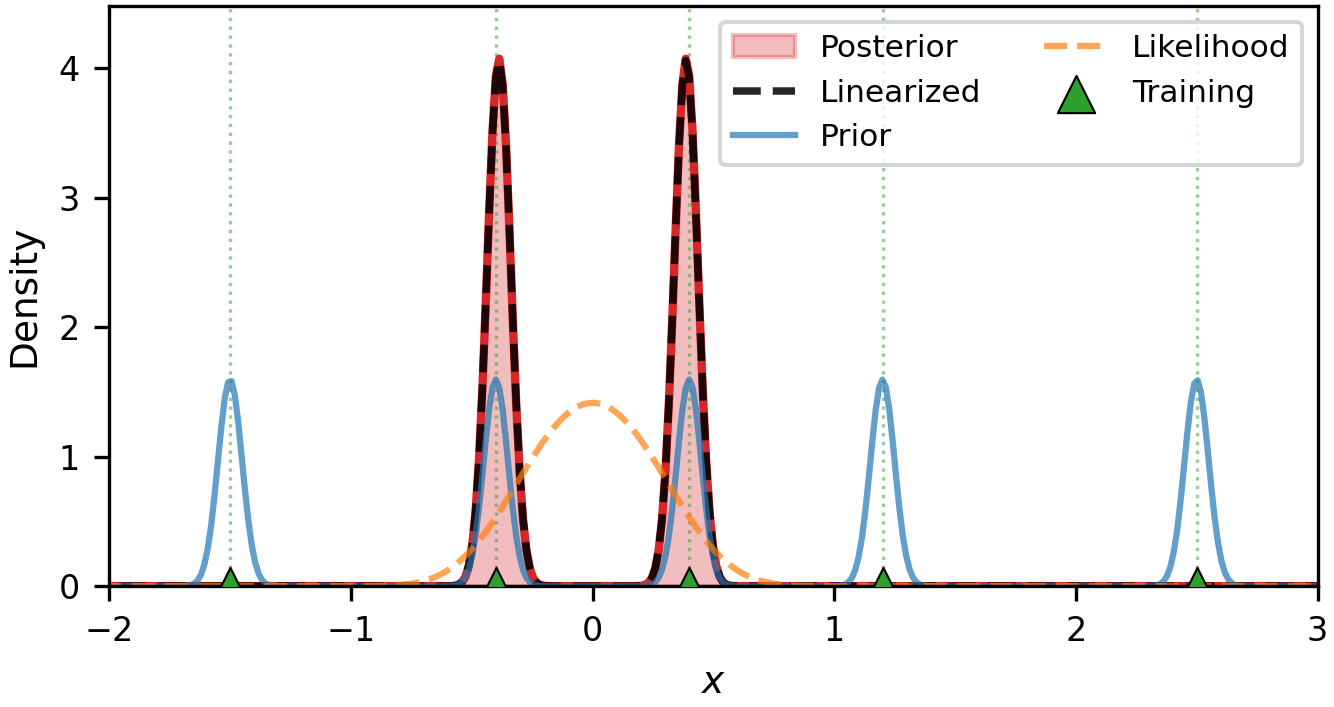} \\[1pt]
\includegraphics[width=0.31\textwidth]{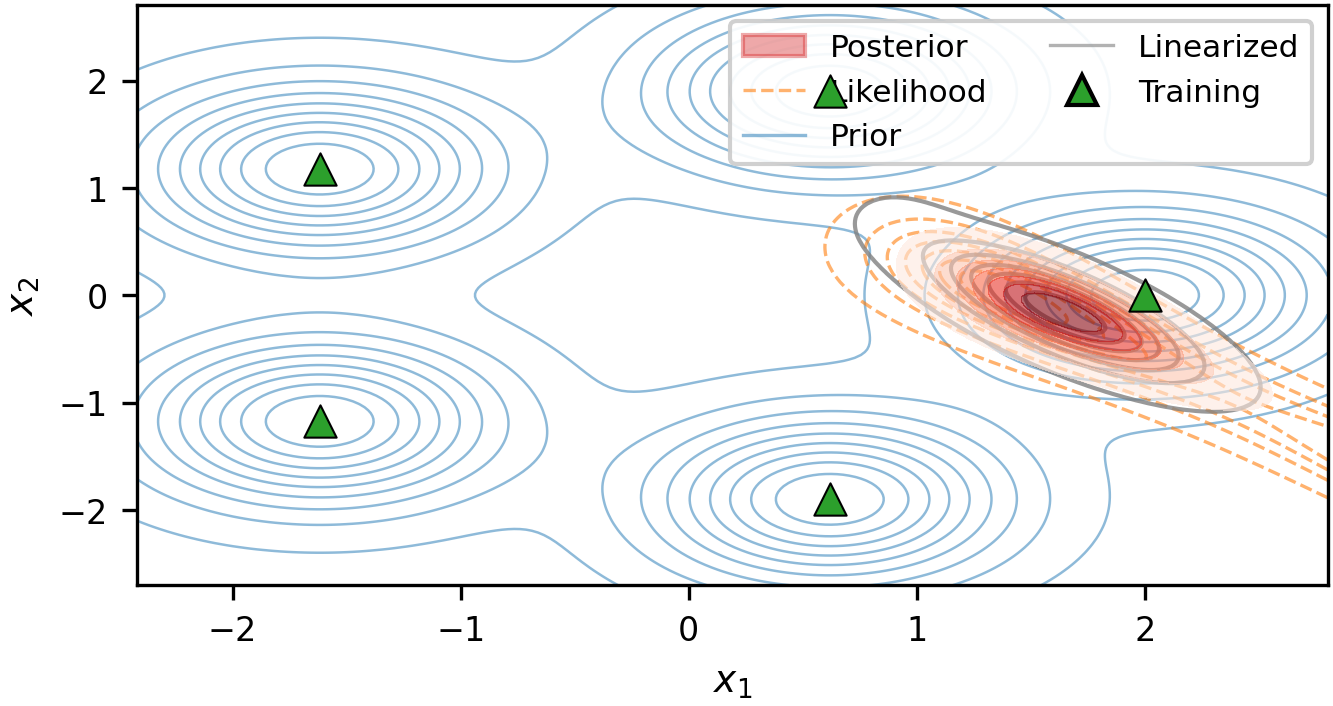} &
\includegraphics[width=0.31\textwidth]{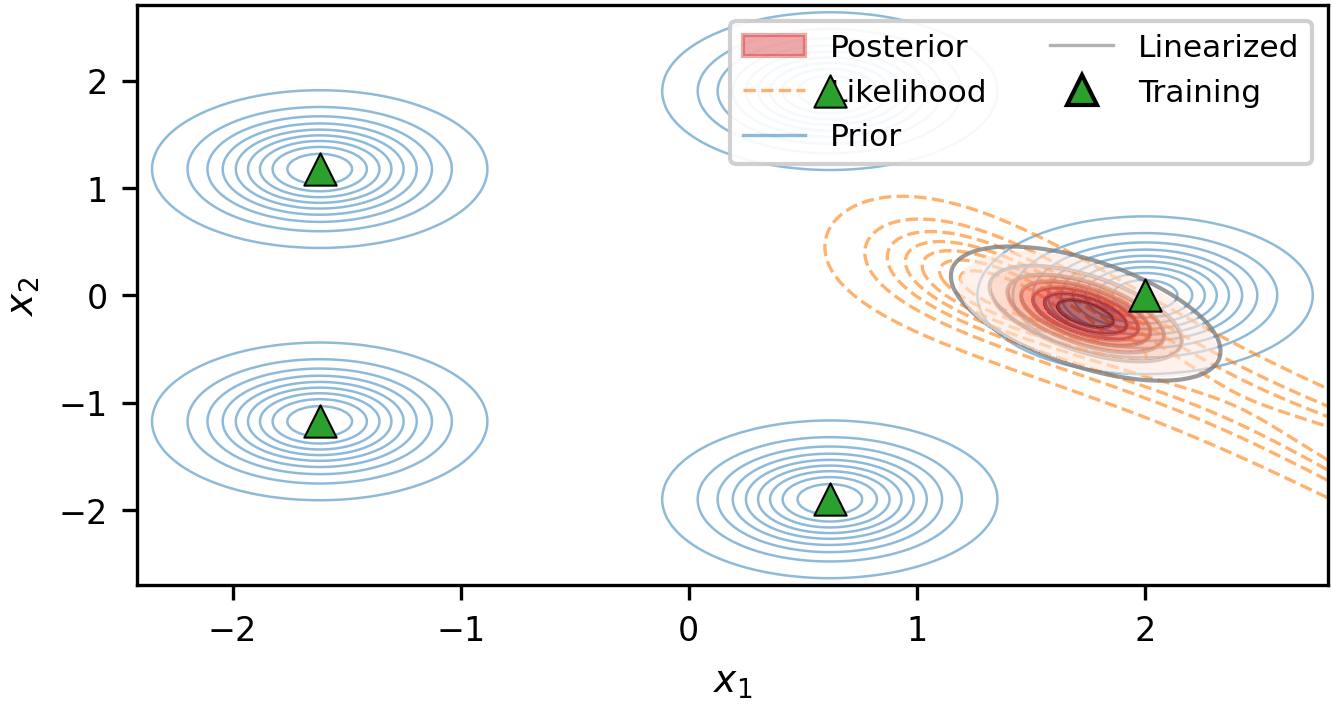} &
\includegraphics[width=0.31\textwidth]{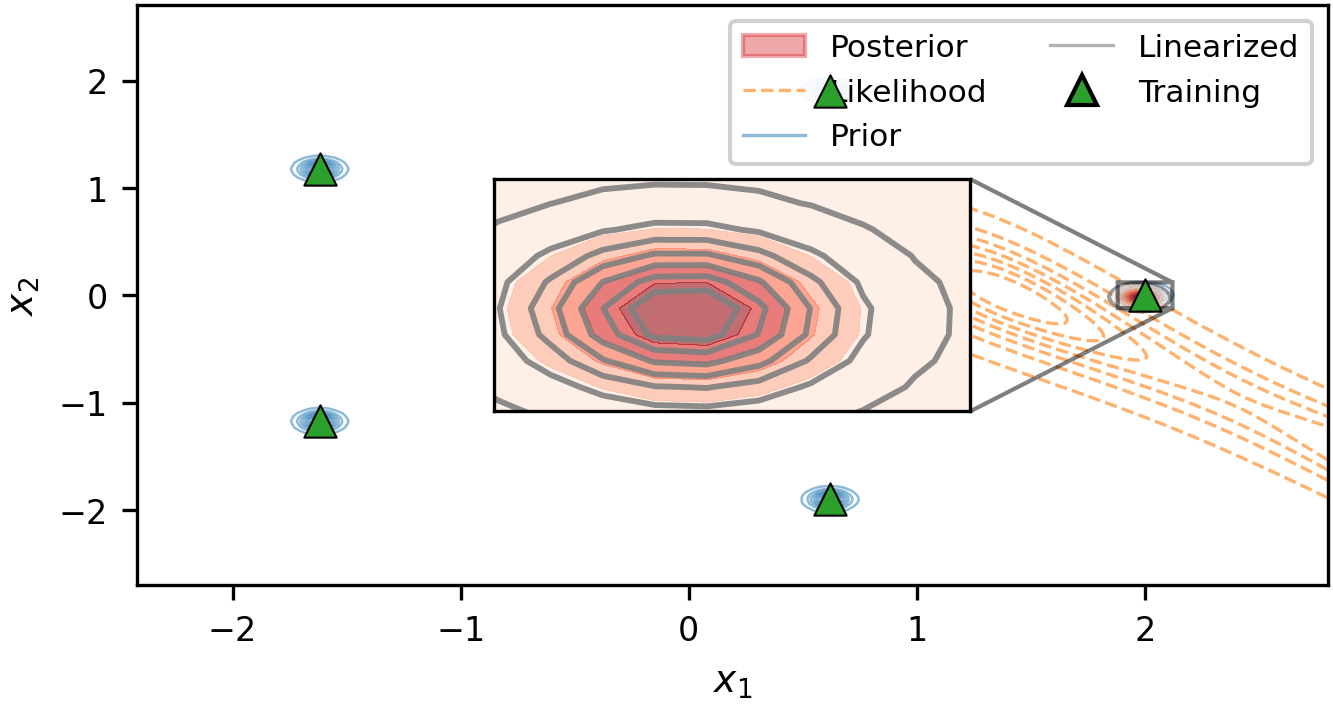} \\
\end{tabular}
\caption{Stylized posterior collapse. Top: 1D posterior (red shaded)
and linearized approximation (black dashed,
Equation~\ref{eq:linearized-posterior}) at $\sigma = 0.5$, $0.3$,
and $0.05$. Bottom: 2D linearized posterior (black contours) overlaid on
exact posterior (red shades). As $\sigma$ decreases, the
posterior collapses from a smooth, data-informed distribution to
discrete spikes at training examples (cf. 
Equation~\ref{eq:lookup-table}).}
\label{fig:stylized}
\end{figure*}

\subsection{Posterior inference under the memorized diffusion prior}

In practice, the reverse process is truncated at a small
$\sigma_{\min} > 0$; we write $\sigma$ for $\sigma_{\min}$ and drop
the scaling $m(t_{\min}) \approx 1$ from the means. It is worth
noting that the analysis below applies at any finite $\sigma$, not
only in the extreme limit of exact convergence to the empirical
distribution. Substituting the
GMM prior, Equation~\ref{eq:gmm-prior}, into Bayes' rule,
Equation~\ref{eq:bayes}, yields a continuous posterior at any
$\sigma > 0$:
\begin{equation}
p(\B{x} \mid \B{y}) \propto
\mathcal{N}\!\left(\B{y} \mid \B{F}(\B{x}),\, \gamma^2 \B{I}\right)
\left[\sum_{n=1}^{N}
\mathcal{N}\!\left(\B{x} \mid \B{x}_n,\, \sigma^2
\B{I}\right)\right].
\label{eq:gmm-posterior}
\end{equation}
Equation~\ref{eq:gmm-posterior} is exact, but the interaction between
each Gaussian prior component and the nonlinear forward operator
$\B{F}$ prevents closed-form evaluation of the posterior.
Because each prior component is concentrated near a training example
(with width $\sigma$), we can linearize the forward operator locally
around each $\B{x}_n$:
$\B{F}(\B{x}) \approx \B{F}(\B{x}_n) + \B{J}_n(\B{x} - \B{x}_n)$,
where $\B{J}_n = \nabla \B{F}(\B{x}_n)$ is the Jacobian evaluated at
$\B{x}_n$. Under the linearized likelihood, each Gaussian prior component
produces a Gaussian posterior component (by completion of the
square), and the full posterior takes the form of a Gaussian
mixture:
\begin{equation}
\begin{split}
p(\B{x} \mid \B{y}) &\approx \sum_{n=1}^{N} w_n\,
\mathcal{N}\!\left(\B{x} \mid \boldsymbol{\mu}_n,\,
\boldsymbol{\Sigma}_n\right), \\
\boldsymbol{\Sigma}_n &= \left(\sigma^{-2}\B{I} +
\gamma^{-2}\B{J}_n^{\!\top} \B{J}_n\right)^{-1}\!, \\
\boldsymbol{\mu}_n &= \B{x}_n + \boldsymbol{\Sigma}_n\,
\gamma^{-2}\B{J}_n^{\!\top}\!\left(\B{y} -
\B{F}(\B{x}_n)\right), \\
w_n &\propto \exp\!\left(-\tfrac{1}{2}
\left\|\B{y} - \B{F}(\B{x}_n)\right\|^2_{
\left(\B{J}_n \sigma^2 \B{J}_n^{\!\top} +
\gamma^2 \B{I}\right)^{-1}}\right).
\end{split}
\label{eq:linearized-posterior}
\end{equation}
In the above expression, $\boldsymbol{\mu}_n$ is the training
example $\B{x}_n$ shifted toward better data fit by
$\boldsymbol{\Sigma}_n \gamma^{-2} \B{J}_n^{\!\top}(\B{y} -
\B{F}(\B{x}_n))$---a correction proportional to the data misfit
projected into model space by the adjoint Jacobian, the same operator
used in gradient-based inversion. The covariance
$\boldsymbol{\Sigma}_n$ balances prior constraint
($\sigma^{-2}\B{I}$) against data constraint
($\gamma^{-2}\B{J}_n^{\!\top}\B{J}_n$): parameters in the null space
of $\B{J}_n$ retain the prior width $\sigma$, while data-sensitive
parameters are narrowed. The weight $w_n$ ranks training examples by
data fit; those with $\B{F}(\B{x}_n)$ far from $\B{y}$ receive
negligible weight.

The linearization is accurate when the curvature of $\B{F}$ over
the $\sigma$-ball around each training example is negligible
compared to $\gamma$. Since each prior component concentrates its
mass within a few $\sigma$ of $\B{x}_n$, the approximation improves
as $\sigma$ decreases---it is most accurate exactly where
memorization is most severe.
As $\sigma \to 0$, the weights $w_n$ reduce to
$\exp(-\|\B{F}(\B{x}_n) - \B{y}\|_2^2 / 2\gamma^2)$, recovering
the lookup-table posterior of Equation~\ref{eq:lookup-table}. In
high dimensions, the volume of each $\sigma$-ball shrinks
exponentially with $d$, so components cease to overlap even at
moderate $\sigma$ and the posterior concentrates on a single
training example.

\section{Numerical illustration}

We validate the theoretical predictions through a stylized
low-dimensional inverse problem---where the GMM prior and linearized
posterior can be evaluated analytically---and through diffusion
posterior sampling for frequency-domain FWI.

\subsection{Stylized example: memorized diffusion model}

We illustrate the theory on two low-dimensional inverse problems where
the GMM prior and linearized posterior can be evaluated in closed form,
using $N = 5$ training examples, observation noise standard deviation
$\gamma = 0.3$, and varying $\sigma$.
The 1D problem uses $F(x) = x + 0.3\,x^3$ with observed data
$y = F(0) = 0$; the 2D problem uses
$\B{F}(\B{x}) = (x_1 + 0.3\,x_1 x_2,\;
x_2 + 0.2\,x_1^2)^{\!\top}$ with training examples at the vertices
of a regular pentagon and $\B{y} = \B{F}(0.7\,\B{x}_1)$---a true
model near but distinct from the nearest training example.
Figure~\ref{fig:stylized} illustrates the transition from
data-informed to memorized posterior as $\sigma$ decreases. At
$\sigma = 0.5$ (left column), the posterior (red) concentrates near
the likelihood (orange), incorporating information from the data. As
$\sigma$ decreases, the prior (blue) sharpens into isolated peaks
at the training examples and dominates the likelihood: at
$\sigma = 0.05$ (right column) the posterior collapses to discrete
spikes at training examples---the lookup table of
Equation~\ref{eq:lookup-table}---ignoring the data entirely. In the
2D case, this collapse pulls the posterior onto the nearest training
example despite the true model being nearby. The linearized
approximation (black dashed in 1D, gray contours in 2D) tracks the
exact posterior with increasing accuracy as $\sigma$ decreases, as
confirmed by the zoomed inset at $\sigma = 0.05$.

\subsection{Frequency-domain FWI with learned diffusion priors}

\begin{figure}[!t]
\centering
\setlength{\tabcolsep}{0pt}
\renewcommand{\arraystretch}{0}
\newcommand{\pw}{0.333\columnwidth}%
\begin{tabular}{@{}c@{}c@{}c@{}}
\includegraphics[width=\pw]{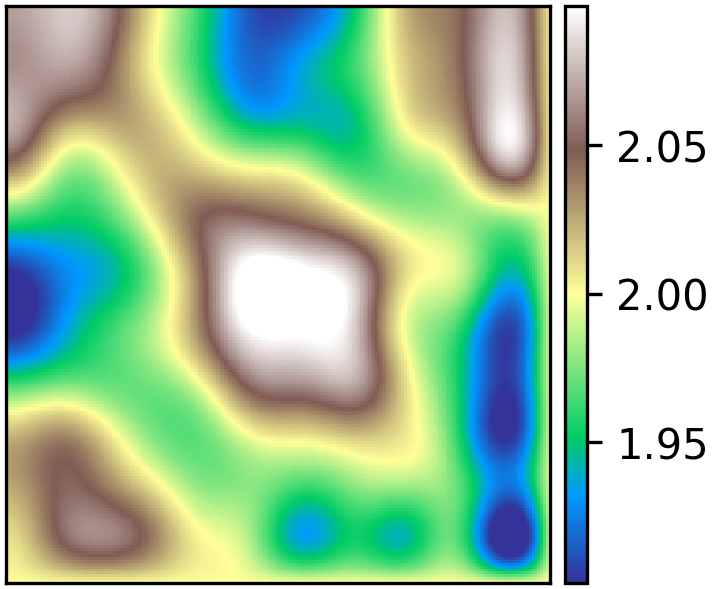} &
\multicolumn{2}{@{}c@{}}{\hspace{4pt}\includegraphics[width=0.64\columnwidth]{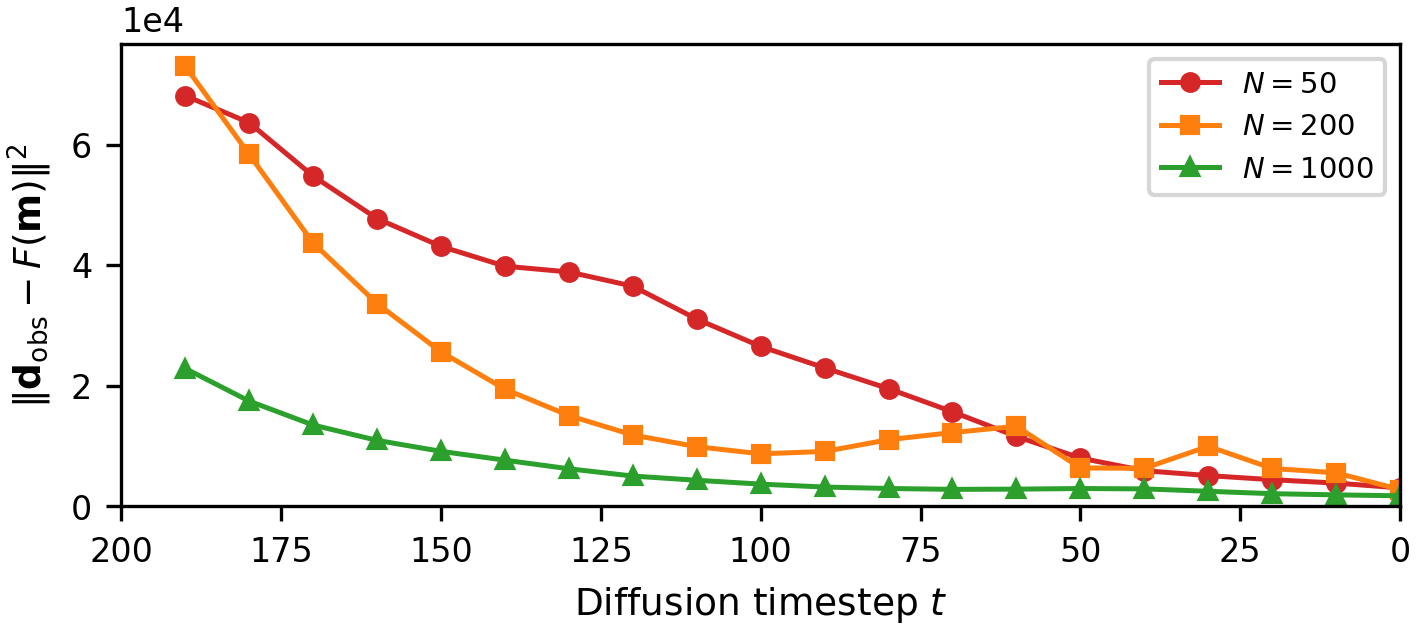}} \\[2pt]
\makebox[\pw]{\scriptsize\textbf{(a)} True model} &
\multicolumn{2}{@{}c@{}}{\scriptsize\textbf{(b)} DPS loss} \\
\end{tabular}
\caption{(a) True velocity model used for the Helmholtz FWI
experiment. (b) Data misfit during the reverse diffusion process
for $N\!=\!50$, $200$, and $1000$.}
\label{fig:helmholtz-setup}
\end{figure}
\begin{figure}[!t]
\centering
\setlength{\tabcolsep}{0pt}
\newcommand{\mpw}{0.25\columnwidth}%
\begin{tabular}{@{}c@{}c@{}c@{}c@{}}
& & \multicolumn{2}{@{}c@{}}{}\\[-4pt]
\makebox[\mpw]{\scriptsize Diffusion sample} &
\makebox[\mpw]{\scriptsize Nearest training} &
\makebox[\mpw]{\scriptsize DPS sample} &
\makebox[\mpw]{\scriptsize Nearest training} \\[2pt]
\includegraphics[width=\mpw]{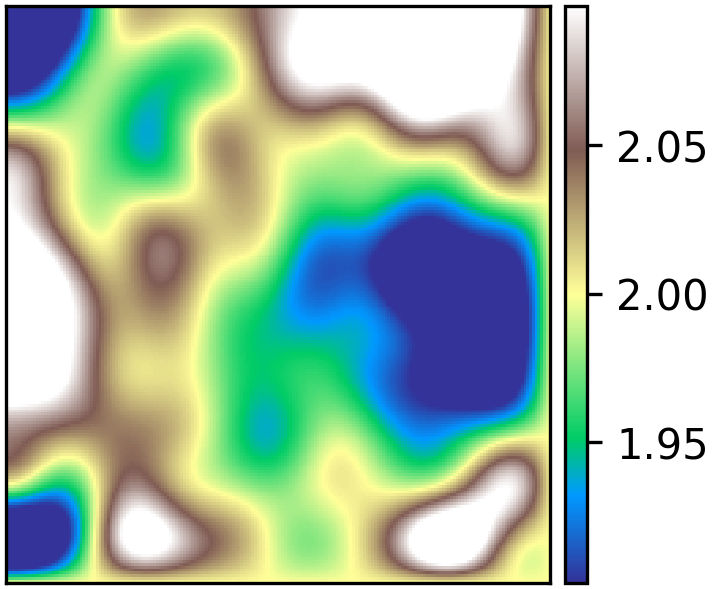} &
\includegraphics[width=\mpw]{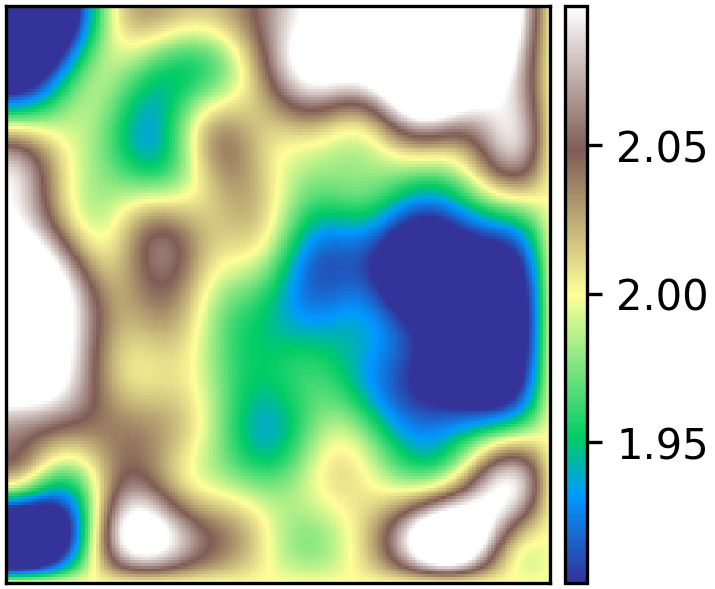} &
\includegraphics[width=\mpw]{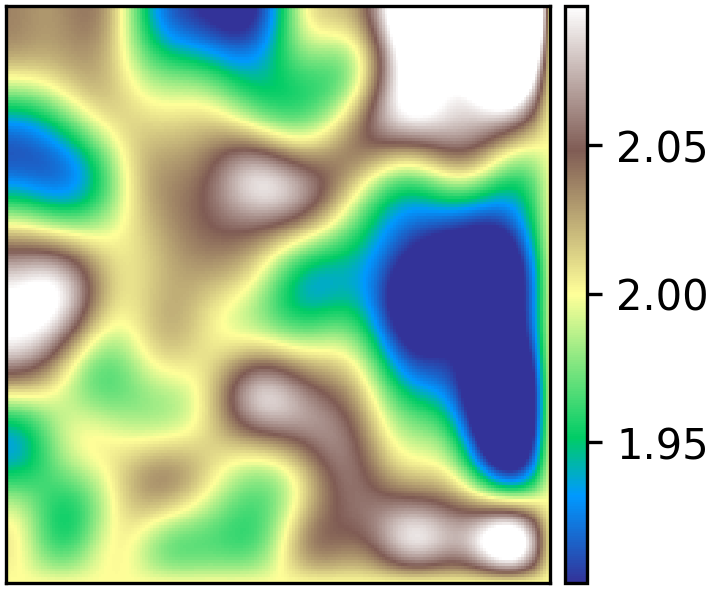} &
\includegraphics[width=\mpw]{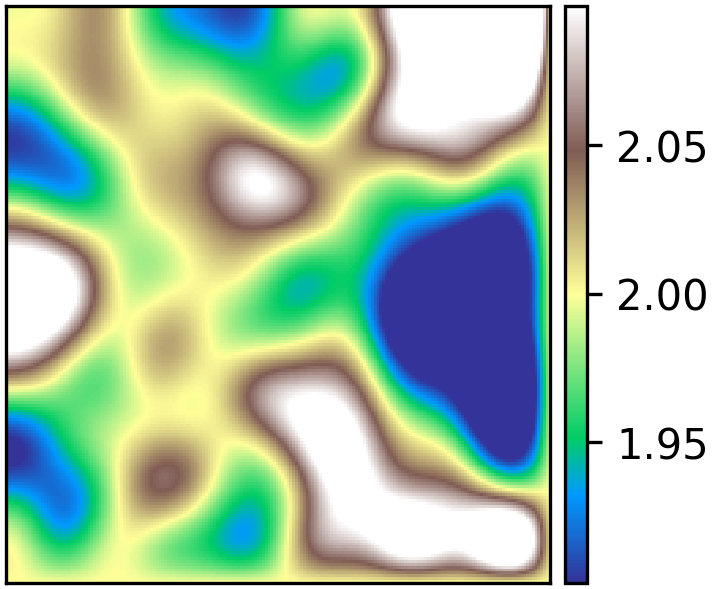} \\[2pt]
\includegraphics[width=\mpw]{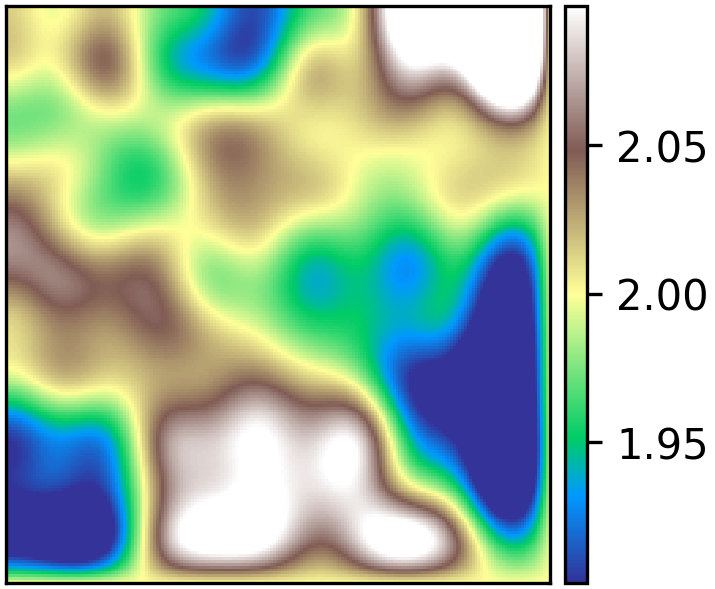} &
\includegraphics[width=\mpw]{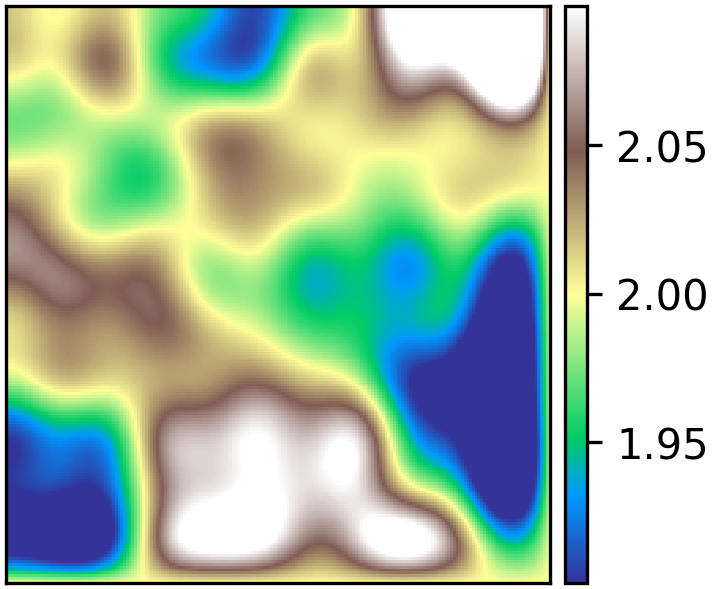} &
\includegraphics[width=\mpw]{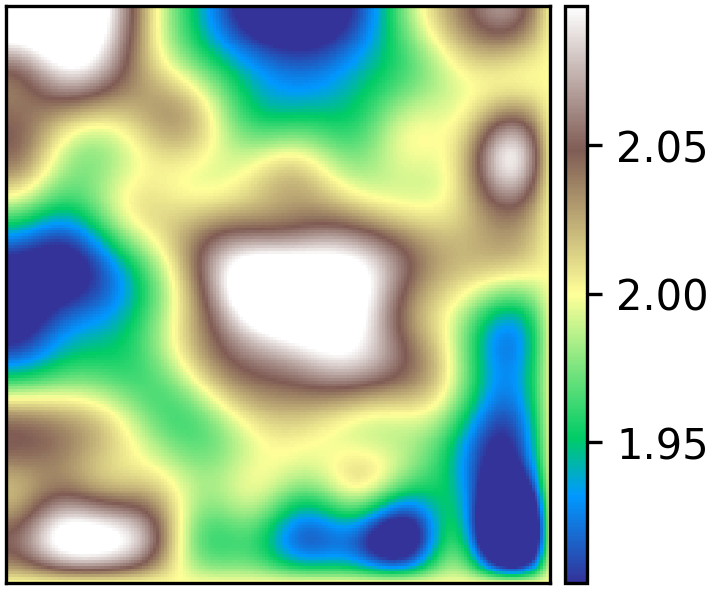} &
\includegraphics[width=\mpw]{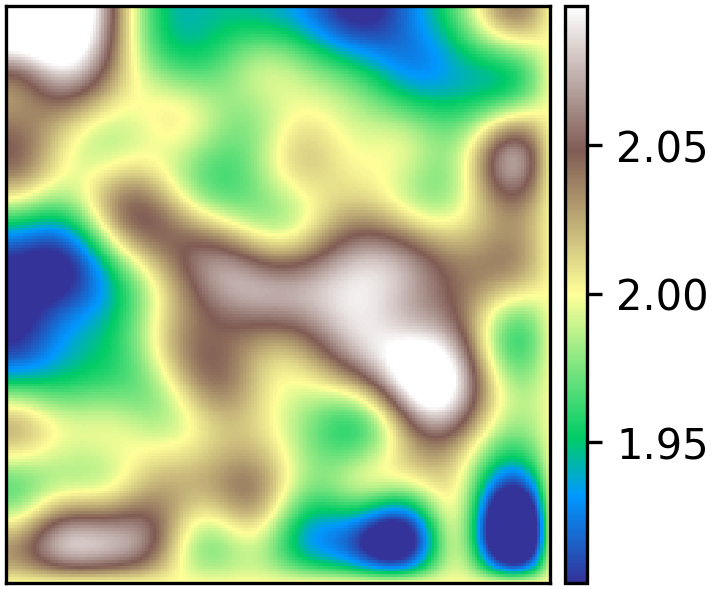} \\[2pt]
\includegraphics[width=\mpw]{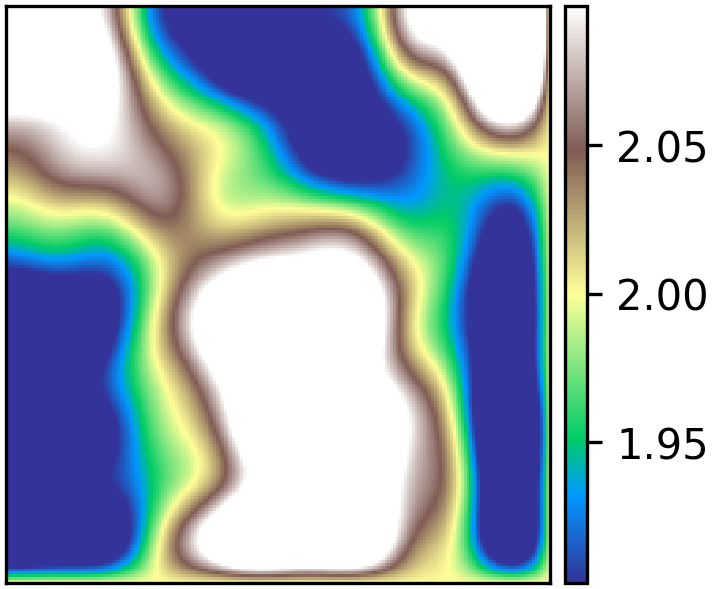} &
\includegraphics[width=\mpw]{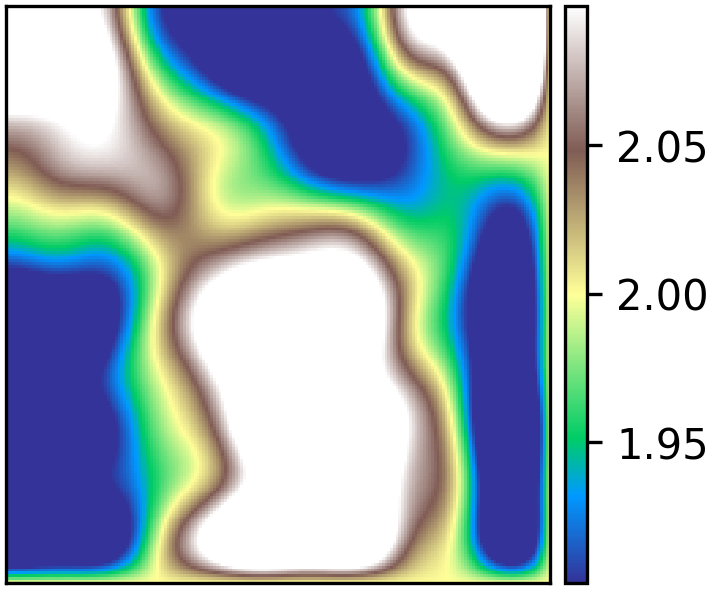} &
\includegraphics[width=\mpw]{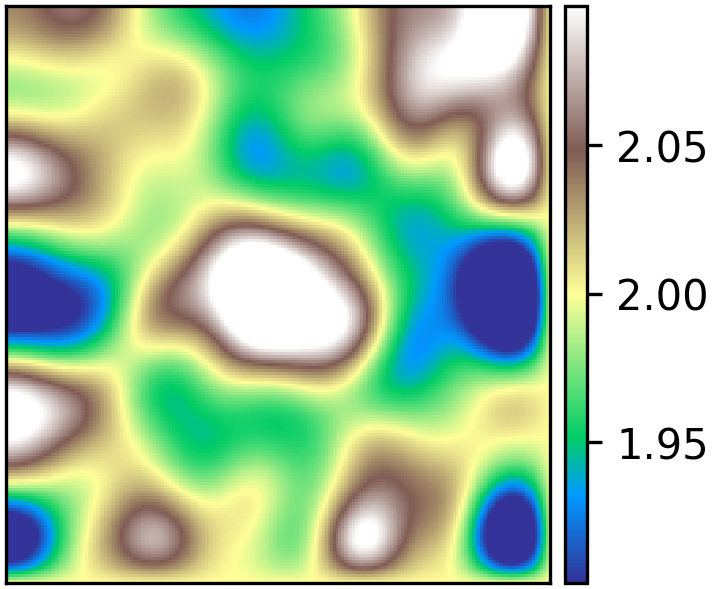} &
\includegraphics[width=\mpw]{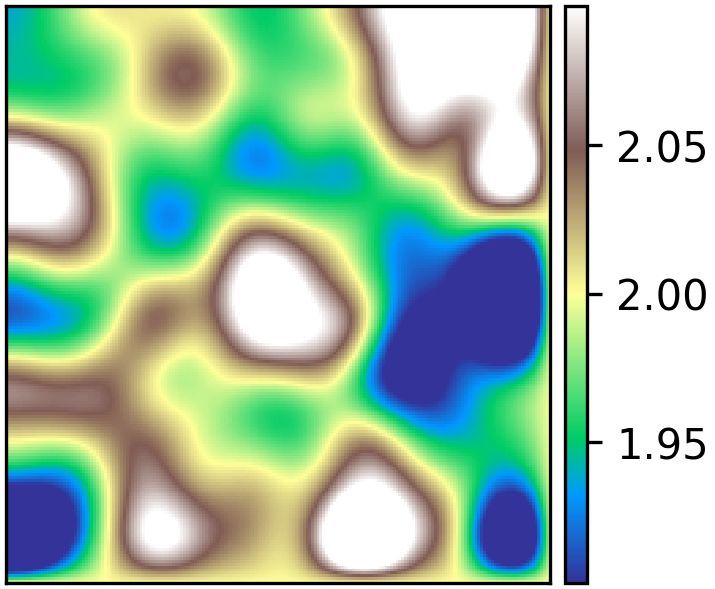} \\
\end{tabular}
\caption{Most memorized $N\!=\!50$ samples and their nearest training
neighbors. Left pair: unconditional diffusion samples are nearly
indistinguishable from training data ($r \approx 0.01$). Right pair:
DPS posterior samples retain memorized structure but the likelihood
introduces perturbations ($r \approx 0.2$--$0.34$).}
\label{fig:memorized-pairs}
\end{figure}
\begin{table}[!t]
\caption{Memorization rates ($r < 0.5$, $256$ samples each) for
unconditional diffusion samples and DPS posterior samples.}
\label{tab:memorization}
\begin{tabular*}{\columnwidth}{@{\extracolsep{\fill}}lccc@{}}
\hline
 & $N\!=\!50$ & $N\!=\!200$ & $N\!=\!1000$ \\
\hline
Diffusion (unconditional) & $100\%$ & $83\%$ & ${{<}1\%}$ \\
DPS posterior         & $58\%$ & $5\%$  & $0\%$ \\
\hline
\end{tabular*}
\end{table}

We test the memorization effect on frequency-domain FWI.
Squared-slowness fields on a $200 \times 200$ grid
($2\,\mathrm{km}$ domain) are expanded in a
$100$-term Karhunen--Lo\`{e}ve (KL) basis derived from the
prior covariance \citep{beskos2017geometric}, reducing the
model dimension to $d = 100$---training on the full
$40{,}000$-dimensional grid would only exacerbate memorization.
MLP-based diffusion models are trained on KL coefficients obtained
by expanding samples from a Gaussian random field (smoothness
$\alpha\!=\!3$, correlation $\tau\!=\!5$) with training sets of size
$N = 50$, $200$, and $1000$,
each iteration-matched at ${\sim}50{,}000$ gradient steps. The
forward operator is the frequency-domain Helmholtz equation at
$6\,\mathrm{Hz}$ with PML absorbing boundaries, $15$ sources at the
surface, and $196$ receivers at the bottom of the domain. The true
model (Figure~\ref{fig:helmholtz-setup}a) is the average of the three most memorized $N\!=\!50$ training
examples in KL coefficient space---close to the training distribution
yet distinct from any individual example. We generate $256$ posterior
samples via diffusion posterior sampling
\citep[DPS,][]{chung2023diffusion} with $5.5\%$ relative observation
noise. The DPS loss curves
(Figure~\ref{fig:helmholtz-setup}b) show that $N\!=\!1000$ achieves
the lowest data misfit, while $N\!=\!50$ retains the highest
residual throughout the reverse process.
Following \citet{wen2024detecting}, a sample is classified as
memorized when its nearest-neighbor ratio
$r = d_1 / \overline{d}_{2:k}$---the distance to the nearest training
example divided by the mean distance to the remaining examples in KL
coefficient space---falls below $0.5$.
Figure~\ref{fig:memorized-pairs} shows the most memorized $N\!=\!50$
samples: unconditional diffusion samples are nearly
indistinguishable from their nearest training examples
($r \approx 0.01$), while DPS posterior samples
show perturbations from the likelihood ($r \approx 0.2$--$0.3$).
Table~\ref{tab:memorization} reports memorization rates. The
$N\!=\!50$ prior memorizes all training data, while
$N\!=\!1000$ shows negligible memorization. The likelihood in DPS
partially corrects the memorized prior---posterior memorization rates
are lower than the corresponding prior rates---but cannot fully
overcome it when the prior is severely memorized.

\begin{figure}[!t]
\centering
\setlength{\tabcolsep}{0pt}
\renewcommand{\arraystretch}{0}
\newlength{\pw}\setlength{\pw}{0.333\columnwidth}%
\begin{tabular}{@{}c@{}c@{}c@{}}
\includegraphics[width=\pw]{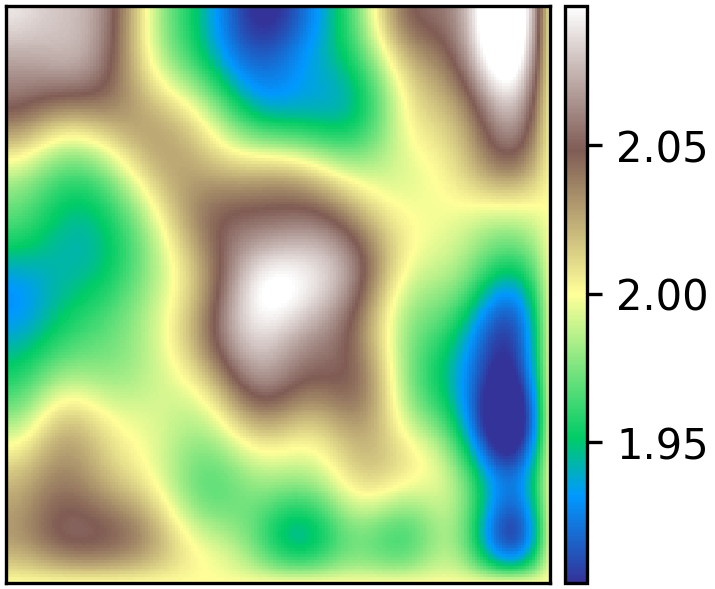} &
\includegraphics[width=\pw]{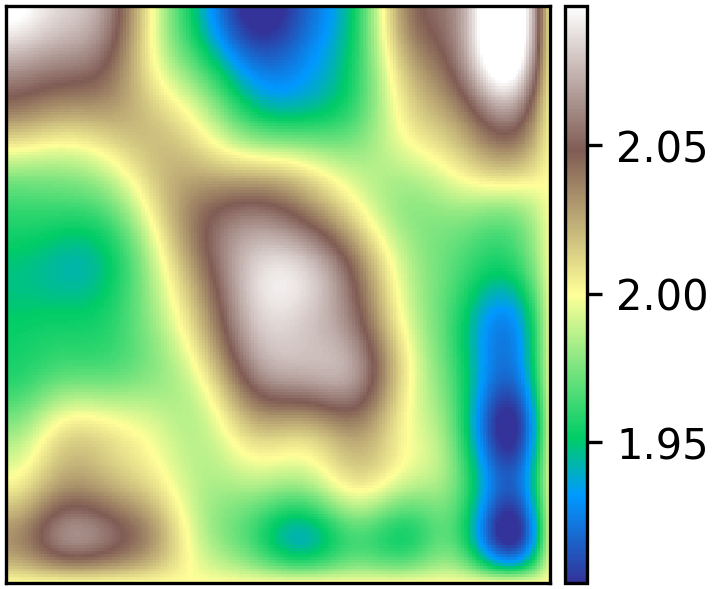} &
\includegraphics[width=\pw]{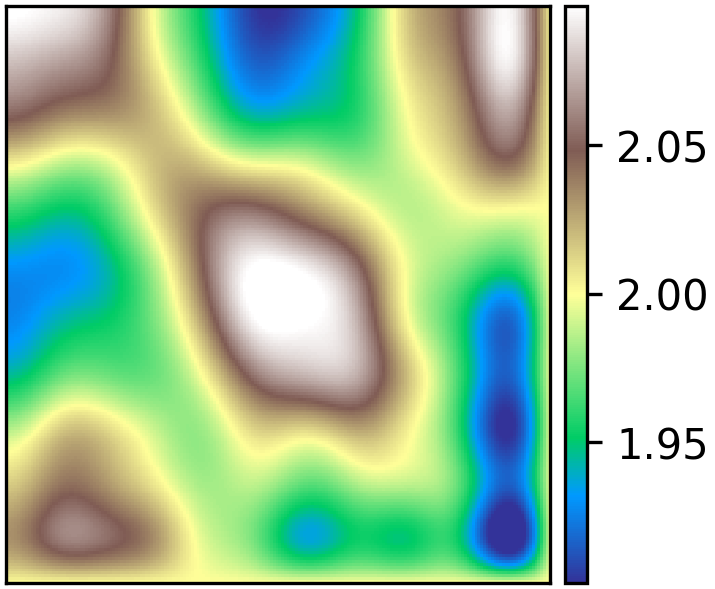} \\[2pt]
\makebox[\pw]{\scriptsize\textbf{(a)} Mean, $N\!=\!50$} &
\makebox[\pw]{\scriptsize\textbf{(b)} Mean, $N\!=\!200$} &
\makebox[\pw]{\scriptsize\textbf{(c)} Mean, $N\!=\!1000$} \\[3pt]
\includegraphics[width=\pw]{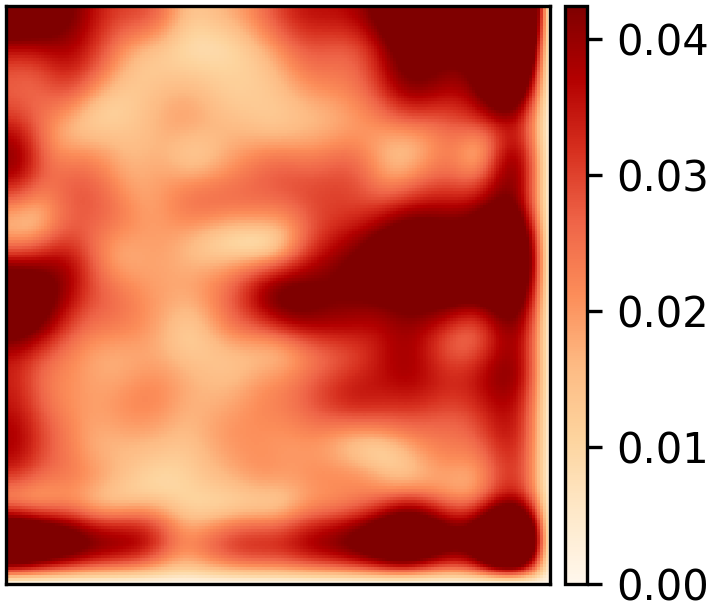} &
\includegraphics[width=\pw]{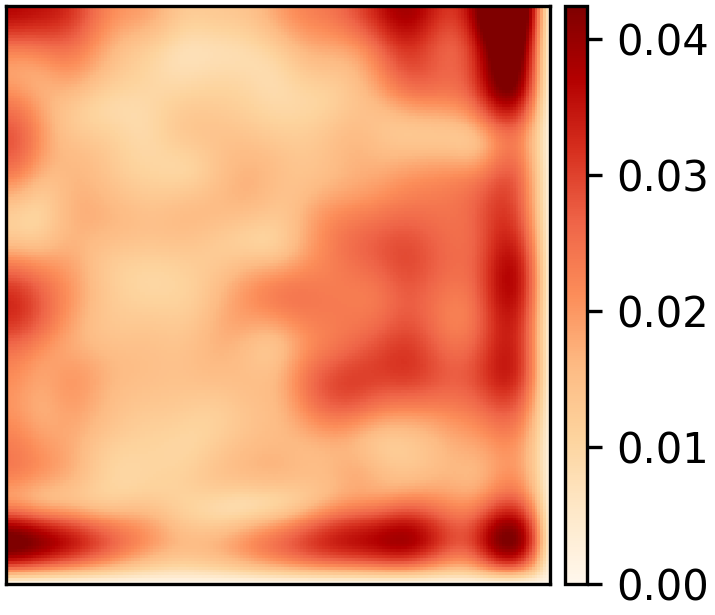} &
\includegraphics[width=\pw]{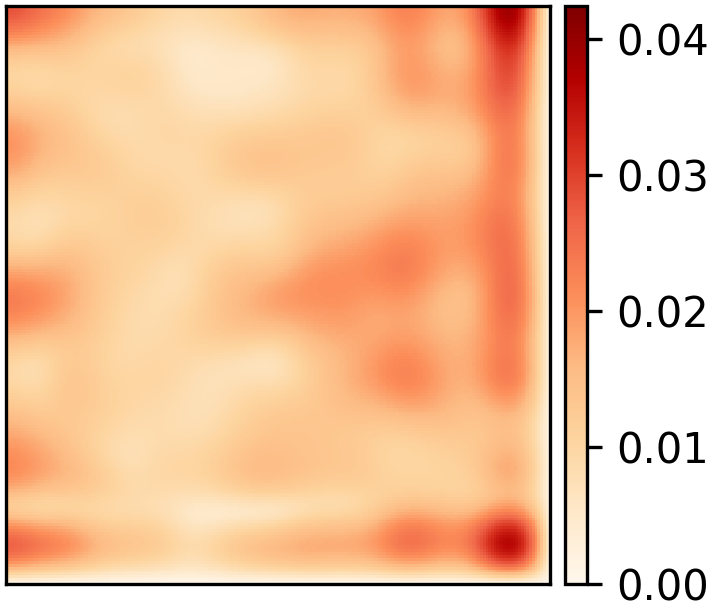} \\[2pt]
\makebox[\pw]{\scriptsize\textbf{(d)} Std, $N\!=\!50$} &
\makebox[\pw]{\scriptsize\textbf{(e)} Std, $N\!=\!200$} &
\makebox[\pw]{\scriptsize\textbf{(f)} Std, $N\!=\!1000$} \\[3pt]
\includegraphics[width=\pw,trim={30pt 0 0 0},clip]{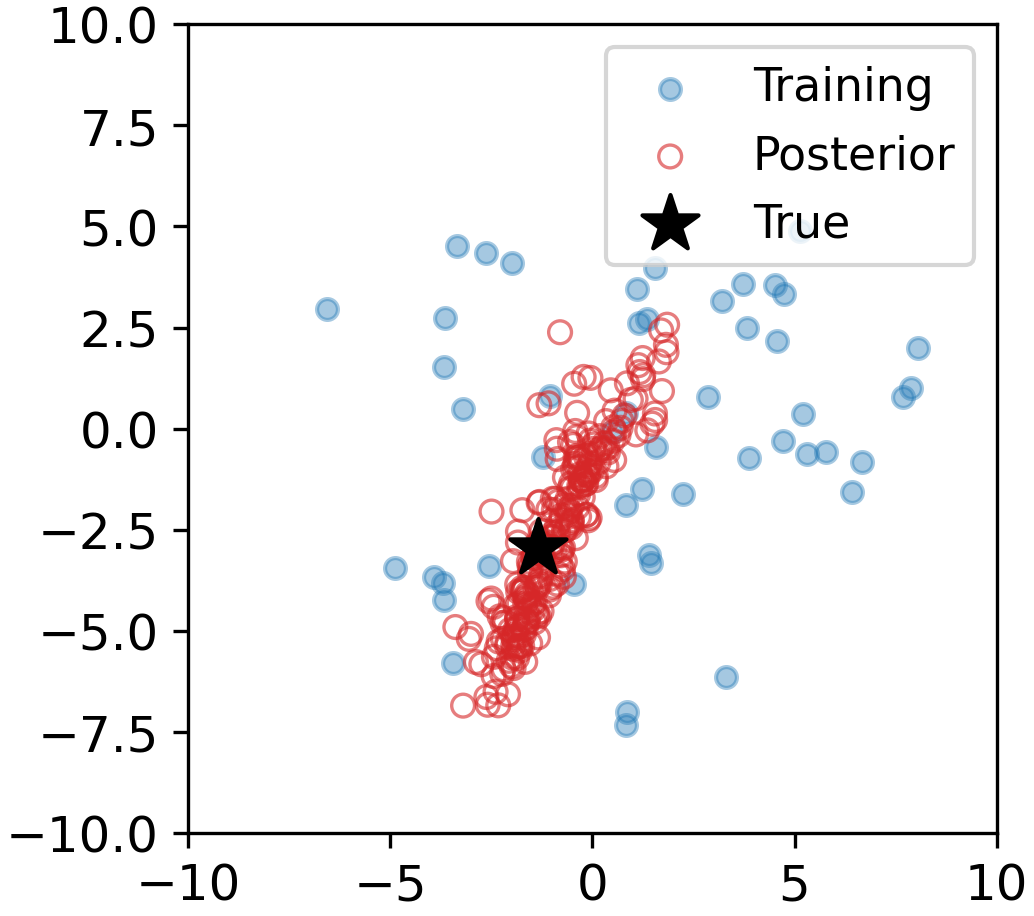} &
\includegraphics[width=\pw,trim={30pt 0 0 0},clip]{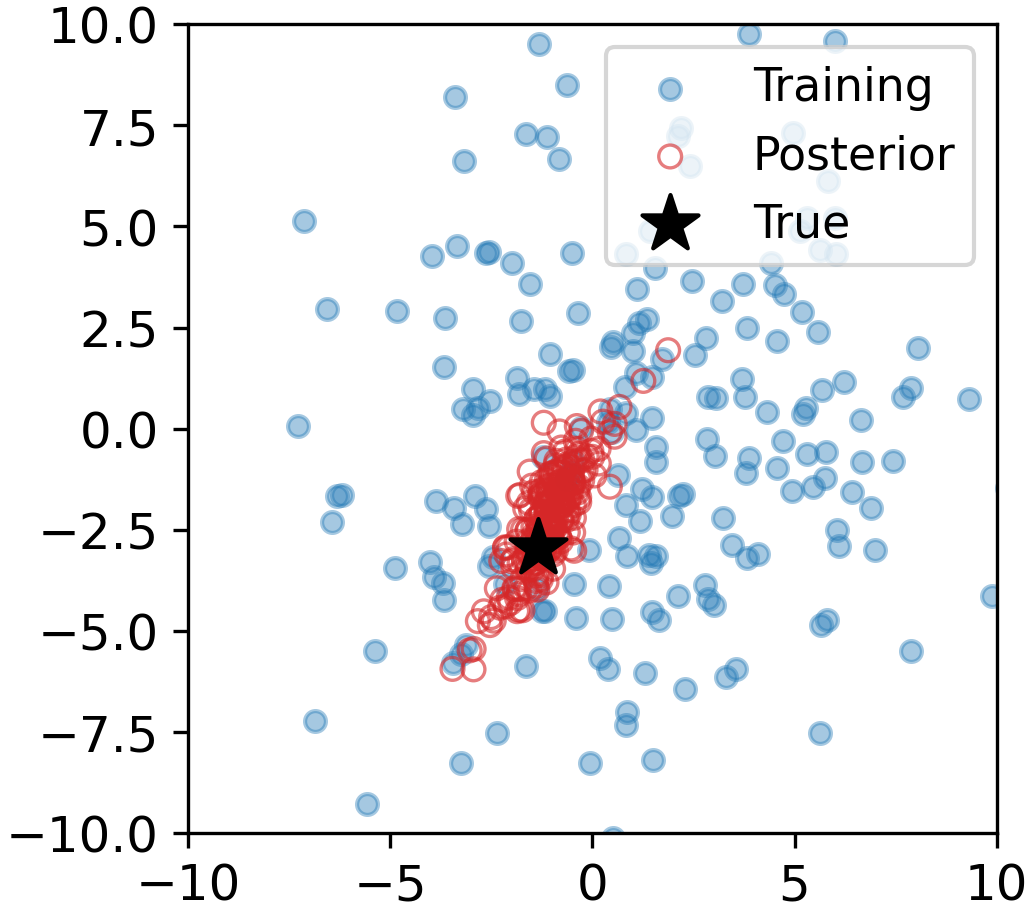} &
\includegraphics[width=\pw,trim={30pt 0 0 0},clip]{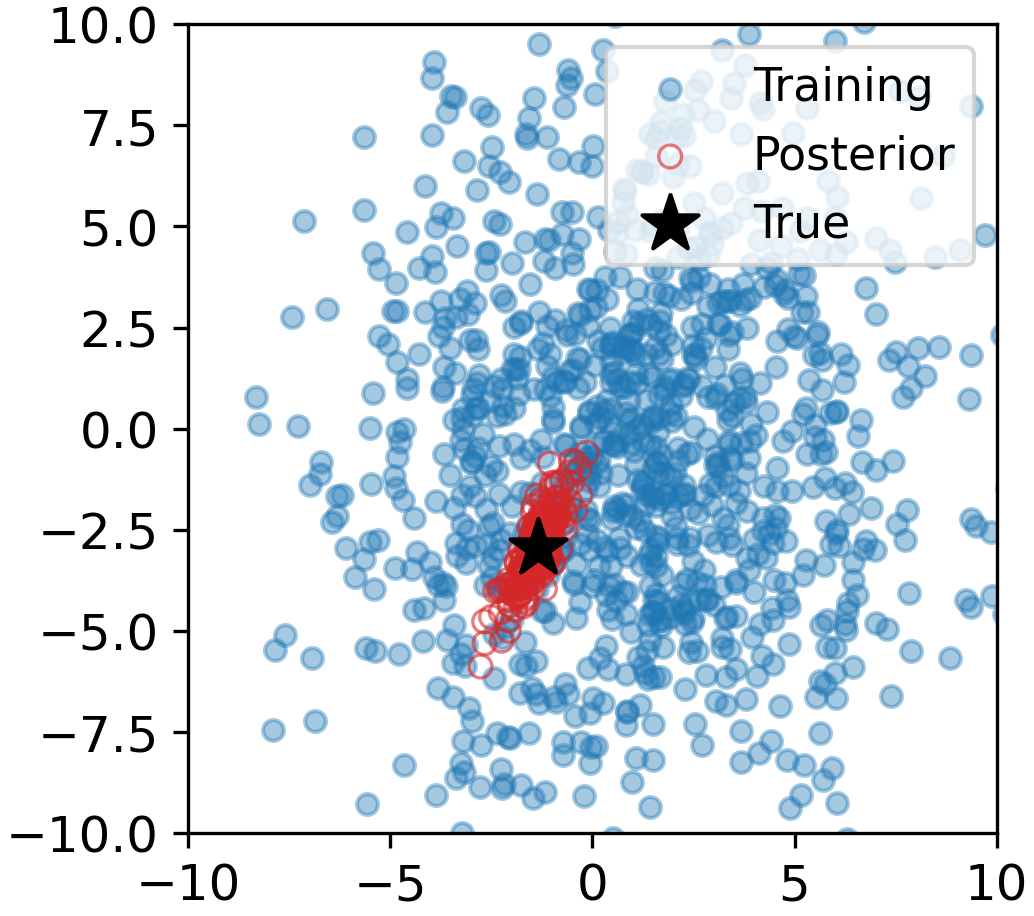} \\[2pt]
\makebox[\pw]{\scriptsize\textbf{(g)} KL scatter, $N\!=\!50$} &
\makebox[\pw]{\scriptsize\textbf{(h)} KL scatter, $N\!=\!200$} &
\makebox[\pw]{\scriptsize\textbf{(i)} KL scatter, $N\!=\!1000$} \\
\end{tabular}
\caption{DPS posterior analysis ($256$ samples, $N\!=\!50$,
$200$, $1000$). (a--c) Posterior mean. (d--f) Pointwise standard
deviation. (g--i) First two KL coefficients: posterior samples
(red), training data (blue), true model (star).}
\label{fig:helmholtz}
\end{figure}

\begin{wrapfigure}{r}{0.46\columnwidth}
\vspace{-2pt}
\centering
\hspace*{-15pt}
\includegraphics[width=0.48\columnwidth]{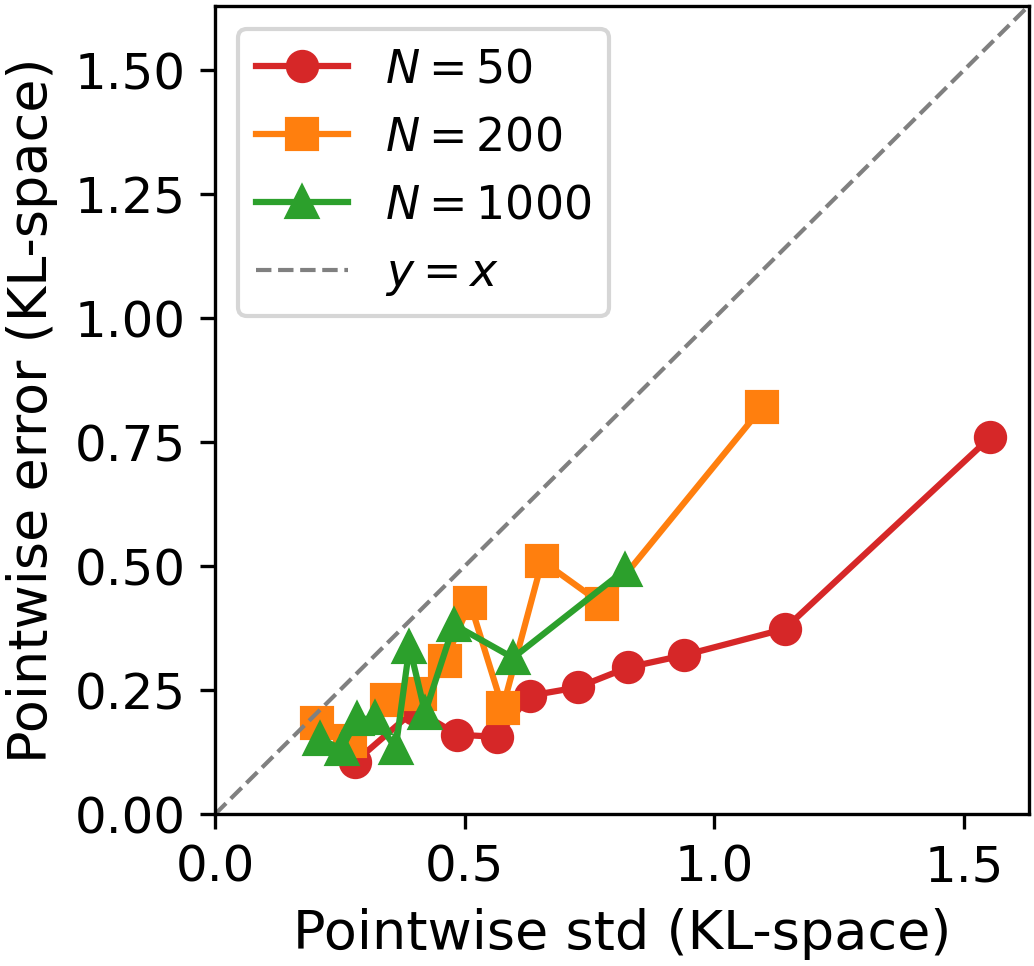}
\vspace{-8pt}
\caption{Calibration: pointwise error vs.\ posterior std in KL space.}
\label{fig:calibration}
\vspace{-12pt}
\end{wrapfigure}
\noindent
The posterior means (Figure~\ref{fig:helmholtz}a--c) improve
accordingly, with $N\!=\!1000$ recovering the true model most
accurately. The pointwise standard deviation
(Figure~\ref{fig:helmholtz}d--f) is largest for $N\!=\!50$, but
as the KL-space scatter plots (Figure~\ref{fig:helmholtz}g--i)
reveal, this spread reflects collapse to different memorized
neighbors rather than genuine posterior uncertainty.
The calibration plot (Figure~\ref{fig:calibration}) confirms this
quantitatively: for $N\!=\!50$, pointwise standard deviation
reaches $1.5$ while error remains below $0.8$---the posterior is
overconfident in its spread rather than in its accuracy. These observations are consistent with the theoretical predictions
of Equations~\ref{eq:lookup-table}
and~\ref{eq:linearized-posterior}.

\section{Discussion and conclusions}

We have shown that when a diffusion model memorizes its training
data, the learned prior reduces to a Gaussian mixture centered on
the training examples, and the resulting posterior becomes a
likelihood-weighted selection among stored models. For diffusion
models specifically, the linearized posterior
(Equation~\ref{eq:linearized-posterior}) reveals how each component
is shifted by the adjoint Jacobian and narrowed by the data---yet
remains anchored to its training example. We do not claim that
learned priors are unsuitable for geoscience; rather, their
deployment requires awareness of the memorization regime, and
memorization diagnostics should become standard practice.

Our analysis focuses on the setting where the true model lies
within the support of the training distribution.
Out-of-distribution models introduce an additional source of
error---the score function itself becomes unreliable far from the
training examples---compounding memorization with biased score
gradients. The same memorization phenomenon arises in conditional
score matching \citep{baptista2025memorization}, implying that
amortized approaches based on conditional diffusion models
\citep{baldassari2024conditional, orozco2025aspire} face analogous
risks. Characterizing the practical consequences of memorization in
these settings, and deriving tight conditions that predict its onset
as a function of training set size, forward operator, and noise
level, remain important open problems. While we used DPS for
demonstration purposes, hybrid physics-based variational inference
methods \citep{siahkoohi2023reliable, orozco2025aspire,
siahkoohi2026dualspace, yin2025wiser}---which ensure physical
consistency and scale more favorably---would be more suitable for
production-scale applications. Potential mitigations include regularization of the score matching
objective \citep{baptista2025memorization} to prevent convergence to
the GMM minimizer, early stopping guided by memorization metrics
\citep{wen2024detecting}, and power-scaling of the prior to restore
the prior--likelihood balance \citep{erdinc2025power}.

\bibliographystyle{seg}
\bibliography{references}

\end{document}